\documentclass[runningheads]{llncs}
\usepackage{graphicx}
\usepackage{tabularx}
\usepackage{answers}
\usepackage{array}
\usepackage{amsmath}
\usepackage{algorithm}
\usepackage{algpseudocode}
\usepackage{tabu}
\usepackage{xparse}
\usepackage{multirow}
\usepackage{amsmath}

\usepackage[utf8]{inputenc} 
\usepackage[T1]{fontenc}    
\usepackage{hyperref}       
\usepackage{url}            
\usepackage{booktabs}       
\usepackage{amsfonts}       
\usepackage{nicefrac}       
\usepackage{microtype}      
\usepackage{lipsum}
\usepackage{longtable}
\usepackage{answers}
\usepackage{algorithm}
\usepackage{xparse}
\usepackage{graphicx}
\usepackage{multirow}
\usepackage{amsmath}
\usepackage{color}

\newcommand{\thickhline}{%
    \noalign {\ifnum 0=`}\fi \hrule height 1pt
    \futurelet \reserved@a \@xhline
}
\newcolumntype{"}{@{\hskip \tabcolsep\vrule width 1pt \tabcolsep}}

\definecolor{purple}{RGB}{150, 0, 255}

\usepackage[normalem]{ulem} 

\begin{document}
\title{Multi-agent Reinforcement Learning for Decentralized Stable Matching}
\titlerunning{MARL for Decentralized Stable Matching}
%
\author{Kshitija Taywade 
\and
Judy Goldsmith 
\and
Brent Harrison
}

\authorrunning{K.~Taywade et al.}
%
\institute{University of Kentucky, Lexington, KY, USA \\
\email{kshitija.taywade@uky.edu}\\
\email{goldsmit@cs.uky.edu}\\
\email{harrison@cs.uky.edu}\\
}
\maketitle              
\begin{abstract}
In the real world, people/entities usually find matches independently and autonomously, such as finding jobs, partners, roommates, etc. It is possible that this search for matches starts with no initial knowledge of the environment. We propose the use of a multi-agent reinforcement learning (MARL) paradigm for a spatially formulated decentralized two-sided matching market with independent and autonomous agents. Having autonomous agents acting independently makes our environment very dynamic and uncertain. Moreover, agents lack the knowledge of preferences of other agents and have to explore the environment and interact with other agents to discover their own preferences through noisy rewards. We think such a setting better approximates the real world and we study the usefulness of our MARL approach for it. Along with conventional stable matching case where agents have strictly ordered preferences, we check the applicability of our approach for stable matching with incomplete lists and ties. We investigate our results for stability, level of instability (for unstable results), and fairness. Our MARL approach mostly yields stable and fair outcomes.

\keywords{Stable Matching  \and Multi-agent Reinforcement Learning \and Decentralized System.}
\end{abstract}

\section{Introduction}
Matching markets are prevalent in the real world, for example, matching of students to colleges, doctors to hospitals, employees to employers, men and women, etc. A two-sided market consists of two disjoint sets of agents. In a two-sided stable matching problem, each participant has preferences over the participants on the other side. A matching is stable if it does not contain a blocking pair. A blocking pair is formed if two agents from disjoint sets prefer each other rather than their current partner. Although, most of the prior literature focuses on centralized algorithms where the entire set of preferences is known to some central agency, having such a central clearinghouse is not always feasible. Therefore, we consider a decentralized matching market with independent and autonomous agents.

There have been several decentralized matching methods proposed in recent years. However, many of them assume that the agents have knowledge of one another's preferences and can easily approach/contact each other, i.e., negligible search friction. In reality, it takes time to meet a partner and to learn the value of said partnership. Furthermore, there is seldom a scope for knowing the preferences of other agents. Also, it is a crucial task to locate and approach a potential match, either by navigating physically or virtually. Several decentralized matching markets, such as worker-employer markets and buyer-seller trading markets, consist of locations at which matching agents may meet, be it physically or online. The level of information, search cost, medium of interaction, and commitment laws can vary across markets. Nonetheless, these are the important features of decentralized markets. Some research works study the impact of these features on the final outcomes for certain types of markets \cite{echenique2012experimental,pais2012decentralized,pais2017decentralized}. To better represent these features, we propose a generalized matching problem in which agents are placed in a grid world environment and must learn to navigate it in order to form matches. We see this as a generalized case for matching problems. While it contains the features described above, it does not conform to the standards set by any individual market type.

There are multiple factors involved in deciding a preferred match in real-world situations, and having a score for each match is more expressive. Thus, we consider weighted preferences for a stable matching problem, which is discussed in \cite{gusfield1987three,irving1987efficient,pini2011stability}. The weighted preference is used as the utility value (or {\em reward}) for being in the match. These scores reflect the underlying preference order. Agents are initially unaware of others' preferences as well as of their own. In many matching markets, knowledge acquisition is important: in labor markets, employers interview workers; in matching markets, men and women date; and in real estate markets, buyers attend open houses. We have taken this into account, so an agent gets to know a noisy version of its utility for a match only after being part of it. Noise represents uncertainty in the value of a partnership, e.g., the uncertain nature of human behavior in relationships.

Finding a long-term match in this scenario is quite a complex task. Therefore, we propose multi-agent reinforcement learning (MARL) as an alternative paradigm where agents must learn how to find a match based on their experiences interacting with others. We equip each agent with their own reinforcement learning (RL) module. We use \emph{SARSA}, a model-free, online RL algorithm. Instead of a common reward signal, each agent has a separate intrinsic reward signal. Therefore, we model this problem as a stochastic/Markov game \cite{littman1994markov}, which is useful in modeling multi-agent decentralized control where the reward function is separate for each agent. Agents learn to operate in the environment with the goal of increasing expected total reward, by getting into a long-term, stable, or close-to-stable and fair match. We impose search cost as a small negative reward (-1) for each step whenever an agent is not in a match.

We investigate the applicability of the MARL approach to the conventional \emph{stable matching (SM)} problem, as well as its extensions, such as \emph{stable matchings with incomplete lists (SMI),} where agents are allowed to declare one or more partners unacceptable \cite{gusfield1989stable}, and \emph{stable matching with ties (SMT),} where agents have the same preference for more than one agent \cite{gusfield1989stable,irving1994stable}. Moreover, we study both the cases of symmetric and asymmetric preferences of agents towards each other. Stability is one of the main measures in our investigation. We check whether our method yields stable results and then to which stable matching the method will converge if there are multiple stable matchings. As we have a dynamic decentralized system with agents having incomplete information, it is hard to guarantee stability for every instance. For unstable outcomes, we check instability with three measures: the degree of instability calculates the number of blocking agents, i.e., those agents who are part of blocking pairs \cite{roth1997turnaround}; the ratio of instability gives the proportion of blocking pairs out of all possible pairs \cite{eriksson2008instability}; and maximum dissatisfaction, which is the maximum difference between an agent's current utility and their obtainable utility by being part of the blocking pair. Overall, we found that many of our outcomes are stable, or if not, they are close-to-stable. Also, it is easy to get stable outcomes for instances with symmetric preferences and harder for the asymmetric ones.

It is important for the outcome to be fair to all the agents, as the goal of the agents is to increase only their own happiness. Therefore, we use three measures of fairness: set-equality cost, regret cost, and egalitarian cost \cite{gusfield1989stable}. We compare the fairness of our results to those of bidirectional local search, a centralized approach, and two decentralized approaches: Hoepman's algorithm \cite{hoepman2004simple}, and a decentralized algorithm by Comola and Fafchamps. Note that these algorithms solve the much easier, non-spatial problems, usually with the assumptions of complete information on the part of agents. Nonetheless, our approach performs competitively in terms of fairness. Lastly, similar to \cite{echenique2012experimental}, we check the proportion of overall median stable matchings, as well as individual median matchings in our results, which are other important measures of fairness.

\section{Related Work}

Reinforcement Learning has not been used for the decentralized two-sided stable matching problem, but researchers have applied both RL and Deep RL mechanisms to solve coalition formation problems. This is closest to our work as matching problems are a special case of coalition formation problems. In \cite{bachrach2018negotiating}, Bachrach et al. proposed a framework for training agents to negotiate and form teams using deep reinforcement learning. They have also formulated the problem spatially. Bayesian reinforcement learning has also been used for coalition formation problems \cite{chalkiadakis2004bayesian,matthews2012competing}. Unlike most of the work in MARL approaches for coalition formation and task allocation, our agents cannot communicate with each other (although they can observe other agents in the same cell). Nonetheless, their utilities get affected by the actions of other agents.

Researchers have studied several decentralized matching markets, and have proposed frameworks for modeling them and techniques for solving them, and have also analyzed different factors that affect the results \cite{niederle2007matching,niederle2009decentralized,satterthwaite2007dynamic,haeringer2011decentralized,echenique2012experimental,pais2012decentralized,pais2017decentralized,diamantoudi2015decentralized}.  Most of these works focus on job markets. Echenique and Yariv, in their study of one-to-one matching markets, proposed a decentralized approach for which stable outcomes are prevalent, but unlike our formulation of a problem, agents have complete information of everyone's preferences \cite{echenique2012experimental}. Unlike our work, none of these works have formulated the problem spatially, and also, they have used different methods than RL. Some distributed algorithms for weighted matching include algorithms that are distributed in terms of agents acting on their own either synchronously or asynchronously \cite{hoepman2004simple,wattenhofer2004distributed,khan2016efficient}. The crucial assumption in these works is that agents already know their preferences over the members of the opposite set and can directly contact other agents to propose matches.

Most of the decentralized methods mentioned here allow agents to make matching offers and accept/reject such offers. While in our approach, an agent shows interest in pairing with an agent from the other set that is present at the same location, by selecting a relevant action. The agent's state space represents those agents from the opposite set that are present at the same cell location and also which ones among them are interested in pairing. Agents get matched only when both the agents select an action for pairing with each other. While this may seem similar to making, accepting, or rejecting offers, it is not exactly the same.

\section{Preliminaries}

Our two-sided stable matching problem consists of $n$ agents divided equally into two disjoint sets $S_1$ and $S_2$. These agents are placed randomly on the grid with dimensions $H \times L$. We investigate if agents can learn good matching policies in a decentralized, spatial setting.

\begin{definition}
In the classical two-sided \textbf{stable matching problem (SM)}, each agent has a strict preference order $p$ over the members of the other set. Given matching $M$, the pair $(i,j)$ with an agent  $i\in S_1$ and an agent $j\in S_2$ is a blocking pair for $M$, if $i$ prefers $j$ and $j$ prefers $i$ to their respective partners in $M$. A matching is said to be stable if it does not contain any blocking pairs.
\end{definition}

The preferences are expressed as weights and hence referred to as weighted preferences. A weight/score represents the true utility value an agent may receive by being in a particular match. These weights still correspond to a strict preference order for each agent. An agent only gets to know the utility from a match when it is in that particular match. Even then, it only receives a noisy utility value for that match rather than the true, underlying utility value. It can be formally written as: for $i\in S_1$ and $j\in S_2$, agent $i$ receives the utility $U_{ij} \cdot C$ for being in a match with $j$, where $C$ is the noise, sampled from a normal distribution with mean $\mu=1$ and standard deviation $\sigma=0.1$ and $U_{ij}$ is the true utility value that agent $i$ can get from a match with agent $j$. Agents still have a strict preference order $p$ over the agents on the other side. $U_{ij}$ is picked uniformly from range $[k,l] \in \mathbb{Z}$, while maintaining the strict preference order.

We also consider following two extensions of the SM problem.

\begin{definition}
The \textbf{stable matching problem with incomplete preference lists (SMI)} may have incomplete preference lists for those involved. In this case, the members of the opposite set who are unacceptable to an agent simply do not appear in their preference list \cite{gusfield1989stable}.
\end{definition}

As we have a score based formulation, an agent has negative scores for unacceptable agents of the other set.

\begin{definition}
Some agents may be indifferent (i.e., have the same utility) between two or more members of the opposite set. This is called the \textbf{stable matching problem with ties (SMT)} \cite{gusfield1989stable,irving1994stable}.
\end{definition}

We consider two types of preferences among agents: symmetric and asymmetric. 

For $i \in S_{1}$ and $j \in S_{2}$, let $p_{i}(j)$ ($p_{j}(i)$, respectively) denote the position of $i$ in $j$’s preference list (the position of $j$ in $i$’s preference list, respectively). In \textbf{symmetric preferences}, $p_{i}(j)=p_{j}(i)$ (in our case, $U_{ij}=U_{ji}$ as well), which is not guaranteed to be the case in \textbf{asymmetric preferences}. With asymmetric preferences (similar to random preferences in literature), there can be many different stable matchings in a market. However, in the case of symmetric preferences, there can be only one stable matching where each agent gets their best choice. Our environment is dynamic and uncertain, and also, due to the narrow difference between noisy utilities, it can be hard for agents to discriminate between their choices efficiently. This can cause unstable outcomes, especially for asymmetric preferences. Therefore, if a stable outcome does not emerge, then we investigate the nature of instability with the following three measures.

\begin{definition}\label{def:DoI}
The \textbf{degree of instability (DoI)} of the matching is the number of blocking agents, i.e., the agents that belong to some blocking pair \cite{roth1997turnaround}.
\end{definition}

Eriksson and Häggström pointed out that, rather than only looking for the number of blocking agents, it can also be helpful to look at the number of blocking partners of an agent, as it gives insight into how likely the agent will exploit instability \cite{eriksson2008instability}. Their notion of instability is defined as follows.

\begin{definition}
For any matching $M$ under preference structure $P^{(m)}$ on a set of $m$ agents, let $B_{P}^{(m)}(M)$ denote the number of blocking pairs. Let $\hat{B}_{P}^{(m)}(M)$ denote the proportion
of blocking pairs:
    $\hat{B}_{P}^{(m)}(M)$ = $B_{P}^{(m)}(M)$ $/$ $m^{2}$
    \cite{eriksson2008instability}.
%
\end{definition}

While Eriksson and Häggström call this measure the `instability' of the matching $M$, we call it the \textbf{ratio of instability (RoI)}. We also use a third measure, \textbf{maximum dissatisfaction (MD)}. It is inspired by the notion of $\alpha$-stability in \cite{pini2011stability} which is specific to SM with weighted preferences.

\begin{definition}
In matching $M$, for every blocking agent $x$, let $y$ be their current match and $v$ be their partner in some blocking pair, then \\
\centerline{
    $\mathit{MD}(M)=\underset{(x,v)}{max} \{ U_{xv} - U_{xy}\}$.
}
\end{definition}

Increase in this number may lead to exploitation of instability by agents in the market. Stability in the outcomes does not guarantee fairness. 
We consider three measures of fairness to check the quality of matchings as given in \cite{gusfield1989stable}.
\begin{definition}
The \textbf{regret cost},
    $r(M)= \underset{(i,j) \in M}{max}$ $max$ $\{p_{i}(j), p_{j}(i)\}.$
\end{definition}

\begin{definition}
The \textbf{egalitarian cost},
    $c(M)= \underset{(i,j) \in M}{\sum} p_{i}(j)$ $+$ $\underset{(i,j) \in M}{\sum} p_{j}(i).$
\end{definition}

\begin{definition}
The \textbf{set-equality cost},
    $d(M)= \underset{(i,j) \in M}{\sum} p_{i}(j)$ $-$ $\underset{(i,j) \in M}{\sum} p_{j}(i).$
\end{definition}

Lower values for these measures indicate better quality of the matchings. Especially, low regret cost and set-equality cost indicate fairness among agents. It is well known that the Gale-Shapley algorithm provides a matching that is optimal for only one side, over all possible stable matchings.  Thus, one notion of fairness is to consider the median of the set of stable matchings, so as to privilege neither set over the other.  Thus, similar to \cite{echenique2012experimental}, we check whether the final matchings are median stable matchings (MSM), and overall, what proportion of individual matches are median matches (MM). The well-known median property is first discovered by Conway \cite{gusfield1989stable}. A median matching exists whenever there is an odd number of stable outcomes. It is the matching that is in the middle of the two sides’ orders of preference. Thus, the median stable matching represents some sense of fairness as it balances the interests of both sides.

\begin{definition}
 
Let $P$ be a preference profile with the set of stable matchings $S(P)$.
If $K=|S(P)|$ is odd, the \textbf{median stable matching (MSM)} is a matching $M \in S(P)$ such that for all agents $a \in S_{1} \cup S_{2}$, $M(a)$ occupies the $\frac{K+1}{2}$th place in $a$'s preference among the agents in $\{M'(a)| M' \in S(P)\}$. $M(a)$ is $a$'s median partner among $a$'s stable-matching partners \cite{echenique2012experimental}. While MSM is for overall matching, \textbf{median match (MM)} refers to individual matches between the agents, i.e., an individual agent being matched to its median stable match partner.
\end{definition}

\begin{center}
   \textbf{Multi-agent Reinforcement Learning}
\end{center}

As mentioned earlier, we propose a multi-agent reinforcement learning (MARL) approach that enables each agent to learn independently to find a good match for itself. A reinforcement learning agent learns by interacting with its environment. The agent perceives the state of the environment and takes an action, which causes the environment to transition into a new state at each time step. The agent receives a reward reflecting the quality of each transition. The agent's goal is to maximize the expected cumulative reward over time \cite{sutton2018reinforcement}. In our system, although agents learn independently and separately, their actions affect the environment and in turn affect the learning process of other agents as well. As agents receive separate intrinsic rewards, we modeled our problem as a Markov game. Stochastic/Markov games \cite{littman1994markov} are used to model multi-agent decentralized control where the reward function is separate for each agent, as each agent works only towards maximizing its own total reward.

A \emph{Markov game} with $n$ players specifies how the state of an environment changes as the result of the joint actions of $n$ players. The game has a finite set of states $S$. The observation function $O$ : $S \times \{1, \ldots, n\} \rightarrow$ $R_{d}$ specifies a $d$-dimensional view of the state space for each player. We write $O_{i} = \{o_{i}| s \in S, o_{i} = O(s, i)\}$ to denote the observation space of player $i$. From each state, players take actions from the set $\{A_{1},\ldots, A_{n}\}$ (one per player). The state changes as a result of the joint action $\langle a_{1}, . . . , a_{n}\rangle \in \langle A_{1}, . . . , A_{n}\rangle$, according to a stochastic transition function $T: S \times A_{1} \times . . .  \times A_{n} \rightarrow$ $\Delta (S)$, where $\Delta (S)$ denotes the set of probability distributions over S. Each player receives an individual reward defined as $r_{i} : S \times A_{1} \times . . .  \times A_{n} \rightarrow \mathbb{R}$ for player $i$. In our multi-agent reinforcement learning approach, each agent learns independently, through its own experience, a behavior policy $\pi_{i}$: $O_{i} \rightarrow$ $\Delta(A_{i})$ (denoted $\pi(a_{i}|o_{i}))$ based on its observation $o_{i}$ and reward $r_{i}$. Each agent's goal is to find policy $\pi_{i}$ which maximizes a long term discounted reward \cite{sutton2018reinforcement}.

\section{Method}

We propose a MARL approach for decentralized two-sided stable matching problems that are formulated spatially on a grid. For each agent, the starting location is picked uniformly randomly from the grid cells. As agents go through episodic training, they start in this same cell location in each episode and explore the environment. Agents must first find each other before they can potentially form matches. This approximates the spatial reality of meeting with individuals (at, e.g., bars or parties) or organizations (at, e.g., job fairs).

We believe that finding a partner for oneself is an independent task, where agents do not necessarily need to compete or even co-operate. Agents only need to learn to find a suitable partner. Each agent independently learns a policy using the RL algorithm, SARSA \cite{rummery1994line,sutton2018reinforcement}, with a multi-layer perceptron as a function approximator to learn the set of Q-values. An agent's learning is independent of other agents' learning as all the agents have separate learning modules (neural networks). We use SARSA because it is an on-policy algorithm in which agents improve on the current policy. Unlike off-policy algorithms like deep Q-learning where agents' behavior while learning can be erratic due to inconsistencies in the policy, on-policy algorithms follow the same policy and improve on it, which is useful when the agent's exploratory behavior matters. In real-world matching markets, there is a value to the path of finding a final match. SARSA is also a model-free algorithm, so that agents directly learn policies, without having to learn the model.

While exploring, agents cannot perceive any part of the environment other than their cell location. If an agent encounters another agent from the opposite set in the same cell and both the agents show interest in
matching with each other, at the same time step, then they get matched. As agents can only view their current grid cell, agents can only match with one another if they are in the same cell. As long as agents are matched, they receive a noisy reward as a utility value at each time step. This noise is sampled from the normal distribution and the true utility value is multiplied by this noise. Note that our environment is deterministic. We now describe the agents' observation space, action space, and reward function.

\vspace{2mm}
\noindent
\textbf{Observation space:} An observation $O_i$ for an agent $i$ at time step $t$, let's say $O_i[t]$, consists of three one-hot vectors. The first one represents an agent's position on the grid, the second vector represents which members of the opposite set are present in the current cell, and third one shows if any of those agents are interested in forming a match. The size of $O_i$ is $R \times C + 2 \cdot m$, where $R$ and $C$ are the number of rows and columns in the grid and $m$ is the total number of members of the opposite set. The size of the first hot vector is equal to the total number of grid cells, and the size of the second and third vector is equal to the size of the opposite set. Thus, an agent initially starts out knowing only the dimensions of the grid and the total number of agents in the opposite set.

\vspace{2mm}
\noindent
\textbf{Action space:}  There are two types of actions available to an agent: navigating the grid and expressing an interest in matching with an agent from the opposite set. The action space is of size $m + 4$, where $m$ is the size of the opposite set and each member has an action associated with it for showing an interest in matching with that member. There are $4$ additional actions for navigating the grid by moving up, down, left, and right. There is no specific action for staying in the same grid cell because whenever an agent is interested in forming a match with another agent, it automatically stays in the same cell. When two collocated agents show an interest in forming a match with one another, then the match is considered to be formed. Note that once a match is formed, the agents must continue to express interest in each other at each time step in order to maintain the match.  If at some point, one ceases to express interest, the match is dissolved.

\vspace{2mm}
\noindent
\textbf{Reward Function:} We have a noisy reward function described as:
$(1)$ $-1$ reward for not being in a match.
$(2)$ The immediate reward received by an agent $i$ for a matching of agents $i$ and $j$ is $R_{ij}=U_{ij} \cdot C$, where $C$ is the noise, sampled from a normal distribution with mean $\mu=1$ and standard deviation $\sigma=0.1$ and $U_{ij}$ is the true utility value that agent $i$ can get from a match with agent $j$.

\vspace{2mm}
Agents have prior knowledge of the grid size and the total number of agents in the opposite set because of the way the states are constructed. However, they completely lack the knowledge of the weighted preferences/utility values of other agents. Furthermore, agents only get to know their own utility for an agent on the other side when they get into a match with it, and that utility value is noisy. In our setup, individuals may choose to be in a match until someone better comes along or may choose to leave a match in order to explore further and look for someone better. Thus, a time step in which all agents are paired is not necessarily \emph{stable}, because agents may break off a partnership to explore, or another, more appealing agent may be willing to partner with them.

\section{Experiments}
 
In this section, we present the ways we tested our approach on stable matching problems. Our main focus is on investigating the applicability of our MARL approach. Along with the classical stable matching case (SM), we examine how MARL performs on variations such as stable matching with incomplete lists (SMI) and ties (SMT). We consider two types of preferences among agents: symmetric and asymmetric. As mentioned earlier, agents have weighted preferences over agents on the other side. For an agent $i\in S_{1}$, it can be seen as the utility value $U_{ij}$ that it gets while in a match with agent $j\in S_{2}$. In the case of SM and SMT problems, these weights are generated from a uniform random distribution in the range $[1,10]$; for SMI, the weights are generated from a uniform random distribution in the range $[-10,10]$ (negative weights indicate how much one agent dislikes the other). For SM and SMI problems, the instances where agents have weights reflecting the strictly ordered preferences are chosen for the experiments. This constraint is removed while choosing SMT instances.

As we formulate the problems on a grid, we investigate results for increasingly complex environments. This complexity is in terms of grid size and the number of agents. 
We use grid sizes $3\times3$, $4\times4$, and $5\times5$ in combination with $8$, $10$, $12$, and $14$ agents as follows:
    $(1)$ Grid: $3\times3$ ; Agents: $8$;
    $(2)$ Grid: $4\times4$ ; Agents: $8$, $10$, $12$, $14$;
    $(3)$ Grid: $5\times5$ ; Agents: $8$, $10$, $12$, $14$.
We do not place more than $8$ agents on a $3\times3$ grid to keep a reasonable density of the population. We chose grid sizes such that agents find other agents easily accessible. This is motivated by real-world places like bars, parties, job fairs, etc. We think that our choices of grid size and number of agents are sufficient to get the essence of realistic situations. Starting cell locations of agents are chosen uniformly randomly from the grid cells, and agents are placed back to these locations at the start of each episode. We run experiments for every possible combination of matching problem variation (SM, SMI, and SMT), preference type (symmetric and asymmetric), grid size, and total agents. We implement $10$ different instances of each of these combinations. Each instance is generated by assigning weights between the agents uniformly randomly while still maintaining the preference order if needed.

\vspace{2mm}
\noindent
\textbf{Parameter Settings: } Each agent independently learns a policy using SARSA \cite{rummery1994line,sutton2018reinforcement} with a multi-layer perceptron as a function approximator to learn a set of Q-values. Each network consists of $2$ hidden layers with $50$ and $25$ hidden units, respectively. We trained models using the Adam optimizer \cite{kingma2014adam} with learning rate $10^{-4}$ to minimize TD-control loss. We used discount factor, $\gamma = 0.9$. We have combined SARSA with experience replay for better results. The use of experience replay along with SARSA has been proposed by Zhao et al. \cite{zhao2016deep}. As SARSA is an on-policy algorithm, we only used data from recent (last 10) episodes in our experience replay buffer, which increased our performance over not using a replay buffer. The number of training episodes and steps varies based on grid-size and the total number of agents in an instance. The number of steps per episode varies between $300$--$700$ and training can take between $100$k to $400$k episodes to converge. When there are multiple suitable matches available in the environment for an agent, a proper exploration strategy is needed to find the best among them. Therefore, we used exploration rate with non-linear decay, such that it is high in the beginning but decays later (with a minimum exploration rate, $\epsilon$ = $0.05$). Learning rate and discount factor are fine tuned as the outcomes are slightly sensitive to these hyper-parameters; however, results are robust to the changes in other hyper-parameters.

\vspace{2mm}
We investigate stability and fairness of the outcomes. Roth hypothesized that the success of a centralized labor market depends on whether the matchmaking mechanism generates a stable matching \cite{roth1991natural}. Although we have decentralized matching market, we think that stability is still an important measure of the success. For the \emph{SM} problem, stable matchings always exist, and for the \emph{SMI} and \emph{SMT} problems, at least a weakly stable matching exists \cite{iwama2008survey}. In weak stability, a blocking pair is defined as $\big(i, j \big)$ such that $M(i) \neq j, j \succ_{i} M(i)$, and $i \succ_{j} M(j)$ \cite{iwama2008survey}. Note that in \emph{SMI} instances, agents can end up without a partner as incomplete lists make some potential matches unacceptable.

As we have a dynamic and uncertain environment and agents with incomplete knowledge, there is a scope for the rise of instability. Economic experiments on decentralized matching markets with incomplete information \cite{unver2005survival,niederle2006making} have yielded outcomes with considerable instability. We use three more measures to study instability: the degree of instability (DoI), the ratio of instability (RoI), and maximum dissatisfaction (MD) (details in Preliminaries). Stable or close-to-stable solutions do not guarantee fairness, specifically for asymmetric preference cases. As agents are independent and autonomous, we need to check the efficacy of our approach from an individual agent's point of view. Therefore, we use three fairness measures: set-equality cost, regret cost, and egalitarian cost. Additionally, we check the proportion of both median stable matchings as well as individual median matches. We compare our results with both centralized and decentralized algorithms. The comparison baselines are detailed below.

\noindent
\textbf{Bidirectional Local Search Algorithm (BLS)} \cite{viet2016bidirectional} is a centralized local search algorithm for
stable matching with set-equality. It uses the Gale-Shapley algorithm \cite{gale1962college} to compute $S_1$-optimal and $S_2$-optimal stable matchings and executes bi-directional search from those matchings until the search frontiers meet.

\noindent
\textbf{Hoepman's Algorithm (HA)} \cite{hoepman2004simple} is a variant of the sequential greedy algorithm \cite{preis1999linear} which computes a weighted matching at most a factor of two away from the maximum. It is a distributed algorithm in which agents asynchronously message each other.

\noindent 
\textbf{Decentralized Algorithm by Comola and Fafchamps (D-CF)} \cite{comola2018experimental} is designed to compute a matching in a decentralized market with deferred acceptance. Deferred acceptance means an agent can be paired with several other partners in the process of reaching their final match. This algorithm includes a sequence of rounds in which agents take turns in making proposals to other agents, who can accept or reject them. While Comola and Fafchamps focused on many-to-many matching, the method can be easily adapted for one-to-one matching. 

Note that not only BLS but also HA and D-CF are non-spatial algorithms where agents already have knowledge of every other agent present in the system. This gives them a significant advantage over the agents in our system, both because the agents know whom they prefer and because they have instantaneous contact, rather than having to wander around in a grid world. Both of the decentralized algorithms use randomness while forming their final matching, giving different results each time. Therefore, we run each instance $5$ times and compare to the average of those runs. We also run our MARL approach $5$ times for each instance. We discovered that if a stable outcome is found, the same one is found consistently, but if not, then the outcomes vary.

\section{Results and Discussion}

We evaluate the results for stability, as well as the level of instability for unstable outcomes. We use three measures to evaluate instability: the degree of instability (DoI), the ratio of instability (RoI), and maximum dissatisfaction (MD). As fairness in the outcomes is also important, we use three fairness measures: set-equality cost, regret cost, and egalitarian cost. In addition to this, we check what percent of the stable matchings are median stable matchings, as well as what percent of the individual matches are median matches. Results for SM and SMT problems with symmetric preferences and for SMI problem with both symmetric and asymmetric preferences are straightforward, therefore, are mentioned in the text. However, the results for SM and SMT problems with asymmetric preferences needed more analysis. We elaborate on the results of SM problem with asymmetric preferences in Tables \ref{tab:table1}--\ref{tab:table3}, as we think that this is the most relevant and adverse case. Due to lack of space, we omitted the similar analysis of the results for the SMT-asymmetric case; however, those results are very similar to the ones presented for the SM-asymmetric case.

{\tiny
\begin{table*}
\setlength{\aboverulesep}{0pt}
\setlength{\belowrulesep}{0pt}
\centering
\begin{tabular}{|p{.66in}|p{.15in}|p{.37in}|p{.37in}|p{.37in}|p{.37in}|p{.37in}|p{.37in}|p{.37in}|p{.37in}|}
\hline
Grid
&\multicolumn{1}{|c|}{$3 \times 3$}
& \multicolumn{4}{|c|}{$4 \times 4$}
& \multicolumn{4}{|c|}{$5 \times 5$}\\
\hline
Agents & $8$ & $8$ & $10$ & $12$ & $14$ & $8$ & $10$ & $12$ & $14$ \\
\hline
Stability(\%) & $100$ & $92.0$ & $82.0$ & $68.0$ & $56.0$ & $80.0$ & $74.0$ & $54.0$ & $46.0$ \\
\hline
DoI & $0$ & $2\pm0.0$ & $2\pm0.0$ & $3.3\pm1.3$ & $3.2\pm1.7$ & $2\pm0.0$ & $2.5\pm1.1$ & $2.8\pm0.9$ & $3.5\pm1.6$ \\
\hline
RoI & $0$ & $0.04\pm0.0$ & $0.04\pm0.0$ & $0.06\pm0.01$ & $0.07\pm0.02$ & $0.04\pm0.0$ & $0.04\pm0.01$ & $0.05\pm0.02$ & $0.07\pm0.03$ \\
\hline 
MD & $0$ & $1.75\pm0.9$ & $2.89\pm1.7$ & $3.13\pm1.9$ & $3.89\pm1.8$ & $2.33\pm1.5$ & $2.77\pm1.4$ & $3.25\pm1.9$ & $4.44\pm2.3$ \\
\hline 
MM(\%) & 83.1 &  73.2  &  65.3 & 63.4  & 52.4 & 75.0 & 67.1 & 58.9  & 48.7\\
\hline
\end{tabular}
\caption{For SM (asymmetric) case, MARL results on stability $(\%)$, instability measures $(Avg\pm Std)$ and median matches $(\%)$.}
\label{tab:table1}
\end{table*}
}

Many of our outcomes are stable or close-to-stable. For SM problem with symmetric preferences, there is only one possible stable matching, and all the outcomes converge to that. However, for asymmetric preferences, more than one stable matching is possible. The instances with symmetric preferences converge faster than the asymmetric ones. The instances of SM and SMT with asymmetric preferences take longer to converge, with lower rates of convergence to stability. The results of SM and SMT are similar. Additionally, for SM asymmetric instances, we have observed that the agents disliked by everyone in the opposite set (low utility associated with them by everyone) find it difficult to get a long-term match. Similarly, unsurprisingly, the most-liked agent (high utility associated with them by everyone) easily settles with its ideal match. We also noticed that the noise in utilities adversely affects convergence to stable outcomes.

When it comes to SMI, our results are always stable. The number of agents that are matched is the maximum possible. This is important because when agents have incomplete lists (negative utilities for matches), it is hard to get a match for everyone, even though it is easier for some agents to find stable partner due to fewer choices. Here, the final outcome always has the lowest regret cost. Importantly, between the agents in the matched pair, there can be an agent having zero utility towards its match, while the other agent still has positive utility for the same match. As the agent with positive utility tries to get in a match, having noise in the reward causes the agent with zero utility to stick to the match. Note that this does not happen when both the agents in the match have zero utility for the match, as neither of them tries to stick with
the match.

{\tiny
\begin{table*}
\setlength{\aboverulesep}{0pt}
\setlength{\belowrulesep}{0pt}
\centering
\begin{tabular}{|p{.12in}|p{.4in}|p{.4in}|p{.35in}|p{.35in}|p{.35in}|p{.4in}|p{.4in}|p{.35in}|p{.35in}|p{.35in}|}
\hline

 \multirow{2}{*}{N}
& \multicolumn{5}{c|}{Set-equality Cost; $d(M)$}

& \multicolumn{5}{c|}{Regret Cost; $r(M)$}\\
\cmidrule{2-11}
 & \shortstack{MARL\\($4\times4$)} & \shortstack{MARL\\($5\times5$)} & BLS & HA & D-CF &  \shortstack{MARL \\($4\times4$)} & \shortstack{MARL\\($5\times5$)} & BLS & HA & D-CF  \\
\hline
 
   $8$ & $3.1\pm2.4$ & $3.9\pm2.5$  & $2.9\pm2.1$ & $2.6\pm1.8$ & $3.1\pm1.7$   & $3.6\pm0.8$ & $3.5\pm0.8$ & $3.5\pm0.8$ & $3.7\pm0.7$ & $3.5\pm0.8$\\
  \hline
  
  $10$ & $2.9\pm2.1$ & $3.2\pm2.8$  & $3\pm2.8$ & $3.5\pm1.6$ &  $4\pm2.7$ & $4.3\pm0.8$ & $4.1\pm0.7$ & $4\pm0.8$ & $4.6\pm0.5$ & $4.2\pm0.9$\\
  \hline
  
   $12$ & $4.6\pm3.5$ & $6.6\pm3.8$  & $5\pm4.2$ & $4\pm4.2$ &  $4\pm4.2$ & $5.4\pm0.8$ & $5.3\pm0.9$ & $5.1\pm0.9$ & $5.3\pm1.1$ & $5.1\pm0.9$\\
  \hline
  
  $14$ & $7\pm9.2$ & $7.4\pm7.7$ & $7.5\pm4.9$ & $7.2\pm4.4$ &  $7.5\pm4.9$ & $6.7\pm0.7$ & $5.9\pm1.3$ & $5.7\pm1.1$ & $5.6\pm1.0$ & $5.7\pm1.1$\\
  
\hline

\end{tabular}
\caption{For SM (asymmetric) case, comparison of set-equality cost and regret cost in $(Avg \pm Std)$ format; results for $8$ agents on $3\times3$ grid not included due to limited space.}

\label{tab:table2}
\end{table*}
}

{\tiny
\begin{table*}
\setlength{\aboverulesep}{0pt}
\setlength{\belowrulesep}{0pt}
\centering

\begin{tabular}{|p{.15in}|p{.6in}|p{.6in}|p{.6in}|p{.6in}|p{.6in}|}
\hline

 \multirow{2}{*}{N}
& \multicolumn{5}{c|}{Egalitarian Cost; $c(M)$}
\\
\cmidrule{2-6}
 & \shortstack{MARL\\($4\times4$)} & \shortstack{MARL\\($5\times5$)} & BLS & HA & D-CF  \\
\hline

   $8$ & $15.3\pm2.8$ & $15.1\pm2.3$ & $15.5\pm2.7$ & $16.6\pm2.8$ & $15.5\pm2.7$  \\
  \hline
  $10$ & $20.3\pm2.7$& $20.4\pm3.4$ & $19.8\pm2.8$ & $25.1\pm3.7$ & $20.4\pm2.9$ \\
  \hline
  $12$ & $31\pm5.9$ & $28.4\pm4.6$ & $27.6\pm3.4$ & $32.6\pm5.1$ & $27.8\pm3.3$ \\
  \hline
  $14$  & $41.4\pm6.1$ & $39.4\pm5.6$ & $34.9\pm3.5$ & $41.2\pm6.9$ & $34.9\pm3.5$  \\
  
\hline

\end{tabular}
\caption{For SM (asymmetric) case, comparison of egalitarian cost in $(Avg \pm Std)$ format.}

\label{tab:table3}
\end{table*}
} 

From Table \ref{tab:table1}, which elaborates on the results of the SM-asymmetric case, we can see that the curse of dimensionality in how the number of agents affects stability. Although the grid size also affects stability, its impact is much less. Both of these factors affect the convergence rate as well: more complex environments take longer to converge. The environment with $8$ agents on a $3\times3$ grid is the easiest one for training agents, and $100\%$ of the outcomes are stable, while the one with $14$ agents on a $5\times5$ grid is the hardest to train and the stability of the final outcomes declined significantly to $46\%$. Nonetheless, we can also see from the measures of instability that the outcomes are close-to-stable. Note that in Table \ref{tab:table1} the values associated with these measures are averaged over only unstable outcomes. The average number of blocking agents (DoI) is low in all cases. We also checked the proportion of blocking pairs (RoI), as the greater this number, the more likely that blocking agents will discover and exploit the instability at some point \cite{eriksson2008instability}. Our approach does well for this measure. This follows the suggestion by Eriksson and Häggström that if agents increase the search effort rather than picking random partners, then we can expect outcomes to have a very small proportion of blocking pairs \cite{eriksson2008instability}.

Furthermore, we look at the maximum dissatisfaction (MD) that an agent can have for an outcome, as great dissatisfaction may also lead to exploiting instability. This number is also low, which assures that there is a low likelihood of blocking agents exploiting unstable outcomes in the market. We think that the dynamic and uncertain environment, incomplete information, noisy utilities, and the narrow differences in the utilities between matches found for an individual over different episodes are potential reasons behind the emergence of instability in the outcomes. Especially in the case of asymmetric preferences, it is unlikely that an agent's ideal partner also best prefers that agent.

In Tables \ref{tab:table2} and \ref{tab:table3}, we compare fairness in the outcomes with three other algorithms. Here, we can see that MARL performs competitively, and there is no significant difference between the fairness results. The regret cost of MARL is slightly, but not significantly, higher for all the types of instances. Hoepman's algorithm (HA) and the decentralized algorithm by Comola and Fafchamps (D-CF) are decentralized approaches. While D-CF always produces stable outcomes, that may not be the case with HA. Our approach performed better than HA in almost all the cases and very similarly to the D-CF algorithm. Again, our approach performs well despite being implemented on a fundamentally more complex formulation of the problem than the ones for HA and D-CF. Further, when our outcomes are stable, they usually match with those found by BLS. It shows that despite the decentralization, our MARL approach is capable of producing outcomes as good as those found by a central agency. This is further supported by the fact that the good proportion of individual matches are median matches (shown in Table \ref{tab:table1}). Also, approximately half of the stable matchings are median stable matchings. We think that the fairness is achieved because agents are self-interested and independent, and the stability is achieved as agents learn to find their best viable matches.

The learned policies include agents moving to a fixed location from their starting point and getting into a match corresponding to the final outcome. It is possible that more than one pair is formed at the same location, but it is rare. The location where agents in a pair move to form the match is not necessarily the mid-point of the distance between starting points of two agents, nor is it guaranteed to be close to either starting point. Centralized algorithms do not work on a grid; they produce matchings but not the learned policies. This shows that the real-world entities can benefit from using our MARL approach to learn to efficiently navigate the environment in finding and maintaining the good match.

\section{Conclusion and Future Work}
We have shown that the MARL paradigm can be successfully used for decentralized stable matching problems that are formulated spatially in a dynamic and uncertain environment, with independent and autonomous agents having minimum initial knowledge. Our MARL approach is also applicable for variations such as SM with incomplete lists and ties. Agents tend to be happy with their final matches, as outcomes are stable or close-to-stable and fair for everyone. Even with unstable outcomes, agents are less likely to exploit instability. In future work, we plan to work on bigger instances, environments where agents can arbitrarily enter and exit the matching market, and investigate environments where a few agents can learn while others have a fixed policy.

\bibliographystyle{splncs04}
\bibliography{references}

\end{document}